# Unicode Normalization and Grapheme Parsing of Indic Languages

***Nazmuddoha Ansary**[1,3], ***Quazi Adibur Rahman Adib**[2,3], **Tahsin Reasat**[3,4],
**Asif Shahriyar Sushmit**[3], **Ahmed Imtiaz Humayun**[3,5], **Sazia Mehnaz**[3],
**Kanij Fatema**[3], **Mohammad Mamun Or Rashid**[6], **Farig Sadeque**[2,3]

[1]Apsis Solutions Limited, [2]Brac University, [3]Bengali.ai, [4]Vanderbilt University,
[5]Rice University, [6]Jahangirnagar University

[1,2,3,6] Bangladesh, [4,5]USA

{nazmuddoha.ansary.28, greasat, ahmed.imtiaz.prio, saziamorshed, fatemakanij52}@gmail.com
quazi.adibur.rahman.adib@g.bracu.ac.bd, sushmit@ieee.org, mamunbd@juniv.edu
farig.sadeque@bracu.ac.bd

## Abstract

Writing systems of Indic languages have orthographic syllables, also known as complex graphemes, as unique horizontal units. A prominent feature of these languages is these complex grapheme units that comprise consonants/consonant conjuncts, vowel diacritics, and consonant diacritics, which, together make a unique Language. Unicode-based writing schemes of these languages often disregard this feature of these languages and encode words as linear sequences of Unicode characters using an intricate scheme of connector characters and font interpreters. Due to this way of using a few dozen Unicode glyphs to write thousands of different unique glyphs (complex graphemes), there are serious ambiguities that lead to malformed words. In this paper, we are proposing two libraries: i) a normalizer for normalizing inconsistencies caused by a Unicode-based encoding scheme for Indic languages and ii) a grapheme parser for Abugida text. It deconstructs words into visually distinct orthographic syllables or complex graphemes and their constituents. Our proposed normalizer is a more efficient and effective tool than the previously used IndicNLP normalizer. Moreover, our parser and normalizer are also suitable tools for general Abugida text processing as they performed well in our robust word-based and NLP experiments. We report the pipeline for the scripts of 7 languages in this work and develop the framework for the integration of more scripts.

## 1. Introduction

Speakers of languages with Alphasyllabary, Akshara, or *Abugida* writing systems (also known as neo-syllabary or pseudo-alphabet) comprise up to 1.3 billion people with a majority being from India, Bangladesh and Thailand. There is a significant academic and commercial interest in developing Natural Language Processing based systems and solutions for these languages. In 1992, Faber suggested "segmentally coded syllabically linear phonographic script" for these languages, while Bright used the term *Alphasyllabary* and Gnanadesikan and Rimzhim, Katz, & Fowler have suggested aksara or ksharik (Gnanadesikan, 2017).

In the Abugida family, the closest analog to 'letters' are the orthographic syllables or the complex graphemes, which have sequential phonemic sequences. These segments act as the smallest written unit in alphasyllabary languages and are termed as *Graphemes* (Fedorova, 2013); the term alphasyllabary itself originates from the alphabet and syllabary qualities of graphemes (Bright, 1999). Each grapheme comprises a *grapheme root*, which can be one character or several char-

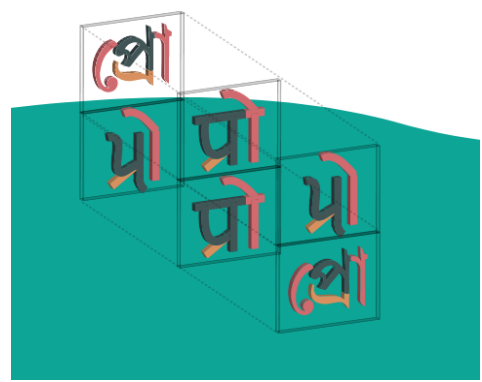

Figure 1: Orthographic components in different languages. Equivalent grapheme constituents are color-coded according to phonemic correspondence. Alpha-syllabary grapheme segments and corresponding characters from the different languages are segregated with the markers. While characters in English are arranged horizontally according to phonemic sequence, the order is not maintained in the Abugida languages.

acters combined as a conjunct. Root characters may be accompanied by *vowel* or *consonant diacritics*– demarcations which correspond to phone-

*These authors share first authorship and contributed equally to this work

mic extensions. To better understand the orthography, we can compare the English word *Proton* to its Indic transliteration প্রোটন [prɔtɔn] (Fig. 1). While in English the characters are horizontally arranged according to phonemic sequence, the first grapheme for both Bangla and Devanagari scripts has a sequence of glyphs that do not correspond to the linear arrangement of Unicode characters or phonemes.

In this paper,

- We discuss a novel orthographic syllable-based encoding scheme for Abugida text and based on it, curated the *Indic Grapheme Parser* which deconstructs words of 7 Indic languages into the different visually sequential complex graphemes. The parser is also easily customizable for any Abugida script.

- We also discuss the existing issues regarding the Unicode encoding of these scripts which hampers both the training and testing performance of natural language processing (NLP) systems, as well as font interpreters, search engines (Sorensen and Roukos, 2007), and other text-based technologies.

- We propose a novel unicode normalizer for the languages to solve the existing issues and compare the normalizer with previously available solutions.

- This work is released under the MIT License. The source code and the experimentation notebooks are available at Github[1,2].

## 2. Characteristics of Abugida Orthography

Abugida words can be deconstructed into visually separable orthographic syllables or graphemes. Each Abugida word is comprised of segmental units called orthographic syllables or graphemes (Kandhadai and Sproat, 2010; Meletis, 2019). The Abugida writing system has language-specific vowel and consonant letters along with vowel and consonant diacritics that are placed up, down, above, below, or around the letters. In almost all cases, vowels have their visually distinct diacritic form. There are also certain consonant diacritics that are visually distinct from the letter it is representing. There are some ligatures that either have i) distinguishable features from the constituent consonants or ii) new conjunct forms. Also, there are certain allographs of both the consonant conjuncts and even the complex graphemes.

[1] https://github.com/mnansary/bnUnicodeNormalizer
[2] https://github.com/mnansary/indicparser

### 2.1. Grapheme Roots and Diacritics

Graphemes in Abugida scripts are constructed by three components: Grapheme Root (vowel, consonant or consonant conjunct), Vowel Diacritic, and Consonant Diacritic. Occurrences of 2 and 3 are optional though both can occur independently in a complex grapheme/orthographic syllable. For Indic languages, these graphemes are often called Akshara. For different Indic languages, placements of these diacritics can be horizontal or vertically adjacent to the root, or surrounding the root. These diacritics may join with the grapheme root and form unique glyphs, from which the constituent roots and diacritics are not visually distinguishable. These also pose the issue of allographs (Sekhar, 2022; Sharma, 2001) in these languages.

### 2.2. Consonant Conjuncts or Ligatures

Consonant conjuncts in Abugida scripts are analogous to ligatures in Latin where two or more consonants are joined to form a glyphs that may or may not visually or phonetically similar than the standalone consonant glyphs. In Bangla, up to four consonants have been found to form consonant conjuncts. A similar is also true for other Indic Abugida writing systems. Although Unicode allows joining even more consonants, font interpreters can only render certain combinations, based on usage. Consonant conjuncts may have two (second order conjuncts, e.g. স্টো [ʃto] = শ [ʃɔ] + ট [t]) or three (third order conjuncts, e.g. ক্ষ্ণ [kʰɔno] = ক [k] + ষ [ʃɔ] + ন [n]) consonants in the cluster. Changes in the order of consonants in a conjunct may result in complete or partial changes in the glyphs. In Table 1 we show some of these consonant conjuncts and the connectors.

### 2.3. Unique Grapheme Combinations

In languages with the Abugida writing system, the graphemes or orthographic syllables are constructed using grapheme roots and vowel and consonant diacritics, which leads to a large number of possible graphemes (Alam et al., 2021). Taking into account the consonants ($n_c$), vowels ($n_v$) and ($n_c^3 + n_c^2$) possible consonant conjuncts (considering 2$^{nd}$ and 3$^{rd}$ order) there can be $((n_c - 3)^3 + (n_c - 3)^2 + (n_c - 3)) + 3 = 3,883,894$ different grapheme roots possible for Bangla. Grapheme roots can have any of the different vowel diacritics ($n_{dv}$) and consonant diacritics ($n_{dc}$). There is only one vowel diacritic in each grapheme, but there can be multiple consonant diacritics in a single one. Although the potential number of possible complex graphemes is quite high, the actual number of complex graphemes in usage has been found to be very low. The number of potential com-plex graphemes may increase in we consider even higher-order conjuncts.

| Script | Connector | Biconsonantal | Tri- and tetra-consonantal conjuncts |
|---|---|---|---|
| Bangla | Hosonto: ্ (2509) | 'ক্স' [kʃɔ] = ক [k] + স [ʃɔ] | ক্ষ্ম [kʰɔ] = ক [k] + ষ [ʃɔ] + ম [m] |
| Devnagari | Halant: ्  (2381) | ज्ह [ɟɦə] = ह [ɦ] + ह [ɦ] | क्ष्ष [kʰkʂ] = ख [kʰ] + क [k] + ष [ʂ] |
| Gurmukhi | Halanta: ੍ (2637) | ਨ੍ਹ [nɦ] = ਨ [n] + ਹ [ɦ] | N/A |
| Gujrati | Halanta: ્ (2765) | સ્થ [stʰə] = સ [s] + થ [tʰ] | N/A |
| Odiya | Halanta: ୍ (2893) | ଵ୍ଖ [wkh] = ଵ [w] + ଖ [kʰa] | N/A |
| Tamil | Pulli: ் (3021) | க்ஷே [kʂeˑ] = க [kə] + ஷ் [ʂ] + ஏ [eˑ] | க்ஷ [kʂə] = க [kə] + ஷ் [ʂ] + ' ' + அ [ə] |
| Malaylam | Chandrakkala: ് (3405) | ന്യ [nja] = ന [n] + യ [j] | ർബ്യ [rətːbja] = ര [r] + ര [r]+ ബ [b] + യ [j] |

Table 1: Examples of bi/tri/tetra-consonantal conjuncts from various Abugida scripts. Most Abugida scripts (such as Devanagari, Gurmukhi, Gujrati, Odiya, Tamil, Malayalam etc.) contain a large number of possible conjuncts which are vastly used but do not have their own Unicode representation. Although *Connectors* are used to circumvent this; it can lead to improper formation of conjunct.

## 3. Characteristics of Abugida Unicode Blocks

The central theme of Unicode-based writing of Abugida scripts is enabling to write numerous complex graphemes with a limited number of Unicode characters, that serve as the smallest constituents. These compact mapping schemes often pose ambiguities by presenting multiple ways of writing visually similar complex graphemes. Also, there are specific complex valid sequences for writing specific complex graphemes. These also pose challenges both for font interpreters (broken render) and also during computational analysis (multiple encoded versions for the same word). Due to the non-unique representation of visual grapheme sequences, different keyboards, and even users following different norms– which leads to multiple variations of the same word, with or without introducing typographic errors.

In Unicode schemes, all the conjuncts are formed as a sequence of consonants joined by a *connector* character. In languages such as Bangla, Tamil, etc. this connector has another role: visually indicating that the consonant does not have an implicit vowel. This dual nature of an essential constituent creates issues with the font interpreters and introduces cognitive spelling errors to an inattentive user. Another prominent feature of Indic Abugida scripts is the *nukta* character (e.g. the bottom dot in Bangla character ৰ). Characters can be written with or without using these separate nukta characters (see Section 4) which causes a large number of non-unique unicode representations as well. As for the conjuncts, vowel diacritics and consonant diacritics can be written freely but only a few specific sequences are allowed. Legacy characters visually similar to standard ones present in the Unicode blocks create even more complications; many keyboards (both mobile and desktop) allow writing these characters– and as many of them look similar to other commonly used characters, unsuspecting users often end up using these characters instead of what they intended. These issues have pushed us towards prioritizing a proper normalization scheme of these texts for their importance in text analysis, language modeling, and other tasks.

In the next section, we categorically discuss the issues with the Unicode representation of prominent Indic languages with Abugida writing systems in detail to demonstrate the issues and the role of the proposed pipeline to correct non-normalized texts.

## 4. Unicode Normalization

### 4.1. Methodology

Our proposed algorithm takes into account the word formation rules using Unicode schemes and fixes seven types of prevalent errors and issues that lead to major non-normalized text formations in Indic Abugida scripts. We also cover some language-specific issues. A previous attempt was made by Kunchukuttan (2020) in making an IndicNLP normalizer; however many of the crucial issues remained unaddressed. Our normalization policy is minimal: going with a less number of Unicode codes to represent glyphs when possible while being consistent. We override the suggestions from the Unicode consortium in cases when this normalization philosophy contradicts them. The issues that are solved by our proposed algorithm are as follows:

#### 4.1.1. Legacy Symbols Handling

These symbols (present in the Unicode blocks) are not used anymore, and some of these look visually similar to widely used characters. Some of the depreciated/legacy symbols for Bangla are ৺, ৭, ৯, ৡ, ২ and for Gurmukhi ੲ, ੳ. These depreciated symbols can cause confusion e.g. ৭ which is known as 'Anji' can be mistaken for ৭ [ʃɛt] (the digit 7) or ৯ [lɪ]which is the character 'Li' can be

17021

mistaken for ৯ [nɔe̯] (the digit 9) for their close visual similarity. Similar is also true for legacy symbols present in other Indic scripts. We provide an option to map these legacy characters to their visually similar counterpart.

### 4.1.2. Broken Vowel Diacritic Fixing

There are multiple occurrences in the Indic Unicode scheme where a combination of two vowel diacritics is visually similar to another one with a separate code; resulting in Unicode ambiguity. For example, in Bangla the combination of ে and া (ো) looks exactly like ো. For the Devanagari script, the combination of ा and ॆ looks exactly like ॊ.

### 4.1.3. Broken Nukta Resolution

The Nukta Unicode issue is a prevalent one in most of the Indic languages. It's a small dot-looking diacritic that gets added with the character to form new ones. There are different nukta signs for different languages. For example, in Bangla usually four characters ড়, য়, র়, ঢ় (IPA: [ɽ, j, ɾ, ɽʰ]) carry nukta, and for all four of them there are two possible Unicode representations: one is the standalone Unicode characters for these, and another one is ড, য, ব, ঢ (IPA: [d, ɟ, b, dʰ]) + nukta and both sets are visually identical with different Unicode representations. The precomposed forms are canonical in standard Unicode for the Bangla nukta issues (Khairullah and Ratul, 2018). In Devanagari, we see the same issue. For example the combination of ज [dʒ] (U+091C Devanagari letter Ja) and ़ (U+093C Devanagari nukta sign) is visually equivalent to ज़ [z] (U+091C Devanagari letter Za). When handling, available normalization produces decomposed forms when using both NFC and NFD. So both approaches are canonically equivalent, but the decomposed form is recommended by the Unicode Standard. This is different from how the nukta issue is handled in Bangla, causing discrepancies in the Unicode processing pipelines. We convert similar characters for all Indic languages into its smallest Unicode sequence form.

### 4.1.4. Invalid Unicode Handling

As we allow a set of certain Unicode to be used in a word, non-glyphs Unicode or Unicode from other languages are cleaned. For instance, the Bangla word অজানা় is not a normalized text; the normalized version of this text is অজানা [ɔɟanɐ]. Previous normalizers sometimes fail to identify this at the ending part of a text.

### 4.1.5. Invalid Connector Handling

There are connectors characters present in Abugida Unicode blocks (example: Hoshonto, Halant, Virama[3] etc.) essentially connects two consonants to form conjuncts (e.g. For Bangla, the connector *hoshonto* carries the code U+09CD). These characters can come at places where they shouldn't (e.g. it can never come between two vowels) and can cause a range of issues. There are certain cases where one may not even realize that this problem is actually caused by an invalid connector. One of the examples of this type is Bangla সং্যুক্তি. The normalized form of this word is সংযুক্তি [ʃɔŋɟukt̪ɪ]. If we obtain words with connector characters present in any place in a sequence where it should not occur (due to typographic or cognitive errors), we fix the word and remove the connector.

In some Bangla texts, we often have to face unwanted middle connectors as well which is also quite hard to identify. For example, another Bangla word চু্ক্তি [cukt̪ɪ] contains an unwanted middle connector. If we normalize it using previously available normalizers and decompose it, we will find [চ [c], ু, ্, ক [k], ্, ত [t̪o], ি]  which is the incorrect form. In our Bangla unicode normalizer, the decomposed output is [চ [c], ু, ক [k], ্, ত [t̪o], ি]– free from any unwanted middle connector (notice the vanished ্ after ু).

We also observed that some Bangla text contains Hoshonto that does not have any language-centric significance. We covered this previously unhandled issue as well. For example, we have seen আমার্ which has an unwanted trailing hoshonto, which we removed (আমার [ɐmɐɾ]).

### 4.1.6. Diacritic Form Correction

In our pipeline, diacritic form corrections include:

- Cleaning consecutive vowel diacritics in unicode: a vowel diacritic can not follow another vowel diacritic.

- Fixing order of consecutive consonant and vowel diacritics: a consonant diacritic can not be followed by a vowel diacritic.

- Cleaning diacritics that follow numbers, punctuations or symbols.

### 4.1.7. Vowel-Vowel Diacritic Removal

A vowel can not be followed by a vowel diacritic. We fix this issue by removing the extra diacritic.

---
[3]https://en.wikipedia.org/wiki/Virama

| Issue | Example | Fix | Fix Visibility | Potential Sources |
|---|---|---|---|---|
| Assamese Replacement (AR) | ব্যরহাৰ [ব, ্, য, ৰ, হ, া, ৰ] | ব্যবহার [bæbohar] [ব, ্, য, ব, হ, া, 'র'] | Closely Resembles | Cognitive |
| Broken Diacritics (BD) | সংস্কৃতি [স, ং, স, ্, ক, ৃ, ত, ি] | সংস্কৃতি [ʃɔŋskr̩tɪ] [স, ং, স, ্, ক, ৃ, ত, ি ] | Little to No | Cognitive |
| Broken Nukta (BN) | কেন্দ্রীয় [ক, ে, ন, ্, দ, ্, র, ী, য়, ্] | কেন্দ্রীয় [kendrɪo] [ক, ে, ন, ্, দ, ্, র, ী, য়] | No. | Keyboard |
| Complex Root Normalization (CRN) | বিষ্পদ [ব, ি, ্, ষ, ্, প, ্, দ] | বিষ্পদ [biʃpod] [ব, ি, ষ, ্, প, দ] | Yes | Cognitive |
| Fix Diacritics (FD) | দুই [দ, ু, ্, ই] | দুই [d̪uɪ] [দ, ু, ই] | No. | Keyboard |
| Invalid Connector (IC) | দুইটি [দ, ু, ্, ই, ্, ট, ি] | দুইটি [d̪uɪtɪ] [দ, ু, ই, ট, ি] | Little to No | Cognitive |
| Invalid Unicode (IU) | াটোবাকো [া, ট, ো, ব, া, ক, ো] | টোবাকো [tobɛko] [ট, ো, ব, া, ক, ো] | Yes | Typo |
| To-hosonto Normalize (THN) | উত্স [উ, ত, ্, স] | উৎস [utʃo] [উ, ৎ, স] | Depends on Font Interpreter | Cognitive |
| Vowel-Vowel Diacritic (VDV) | একএে [এ, ক, এ, ে] | একত্রে [ekotre] [এ, ক, ত, ্, র, ে] | Yes | Typo and Cognitive |

Table 2: Major Unicode Issues in Bangla with Examples. Does not include legacy character issues. CRN and THN are Bangla-specific, all others are applicable to other writing systems.

### 4.1.8. Language Specific Treatment

In addition to these seven, we provide four more operations to be used specifically for Bangla. They are:

**Unwanted doubles** Due to mistyping or keyboard issues, unwanted double diacritics can be found in a Bangla text. For instance, normalized form of যুদ্ধ is যুদ্ধ [jud̪d̪ʰo]. It is really hard to distinguish these words– but they are, as we will see, and our proposed normalizer can solve this issue. The decomposed version of a normalized form is [য [jo], ু, দ [d̪], ্, ধ [d̪ʰɔ]]. Previous efforts are unable to handle this type of case: their version of the decomposed form is [য [jo], ু, ্, দ [d̪], ্, ধ [d̪ʰɔ]], which, unfortunately, is not correct.

**Complex root normalization** Complex grapheme roots are normalized by allowing popular combination of connected Unicode. For example, in Bangla language কককক্ক is a valid combination in terms of Unicode but this combination is never used in real life so the allowed form 'ক্ক' [kk] is provided as the normalized form. Randomized connectors are also normalized through complex roots such as বিষ্পদ = ব+ি+্+ষ+্+প+্+দ, which is normalized as বিষ্পদ [biʃpɔd] = ব[b]+ি+ষ[ʃ]+্+প [po]+দ[d̪] in Bangla language.

Grammatically, It is impractical to get multiple Fola (curtailed consonants) in a row. For instance, in Bangla language ন্ব্ৰ does not carry any linguistic value whatsoever, yet it is a writable grapheme. So, It is necessary to remove this and provide a normalized form. Our developed normalizer is able to handle this case which was ignored previously.

Our normalizer can normalize even more complex roots which were not normalized by previous efforts. For example, for Bangla আকাক্ষ্মা's normalized form is আকাঙ্ক্ষা [ɛkɛɛɛŋʃɛ] but it was normalized as আকাক্ক্ষা which is inaccurate. A combination of multiple conjunct diacritics that requires a language-specific order is addressed in our pipeline.

**Assamese character replacement for Bangla** Assamese and Bangla both share the Eastern Nagri script and hence have the same Unicode block. Sometimes characters are found in Bangla text that closely resembles Bangla but is in fact Assamese. We cleaned these characters from the Bangla scripts.

**To-hoshonto normalization**
ৎ [æ] symbol which is formed from ত [to] and ্ and no space char (u200c) should be replaced with ৎ (Constable, 2004). We normalized [ত [to], ্] as ৎ [æ]. This structure holds the actual sense of ৎ [æ] and was not previously done.

Table 2 shows the explanations of the different normalization schemes. Of these, 4 are specific to Bangla and the others are general and valid for any Abugida script written on unicode-based keyboards.

## 4.2. Word Level Experiments

We tested the effectiveness of our normalizer on Indic texts collected from the Internet. We ran our normalizer on words collected from the OSCAR online multilingual corpus (Abadji et al., 2022). This corpus contains texts from diverse online domains such as blogs, newspapers, social media, etc. thus containing commonly occurring issues in digitally written texts using widely available keyboards for seven Indic Abugida languages. Table 3 contains some basic information on this corpus; for example, the corpus had 6,885,008 unique Punjabi words, and 214,242 of those were subjected to modification by the normalizer, which is

| Language | Total | Affected | % |
|---|---|---|---|
| Bangla | 2,883,731 | 369,348 | 12.81 |
| Devanagari | 2,887,725 | 257,615 | 8.92 |
| Gujarati | 1,119,927 | 28,512 | 2.55 |
| Odiya | 417,483 | 36,267 | 8.69 |
| Tamil | 6,885,008 | 214,242 | 3.11 |
| Punjabi | 421,537 | 132,993 | 31.55 |
| Malayalam | 6,021,714 | 932,314 | 15.48 |

Table 3: Summary of the OSCAR Abugida language corpus. It shows how many, out of the total unique words, our pipeline was able to capture as unnormalized words

3.11% of the Punjabi words, written using the Gurmukhi script.

### 4.3. NLP Experiments

We compare our proposed normalizer with IndicNLP (Kunchukuttan, 2020). We ran experiments on a Bangla multi-class NER (Ashrafi et al., 2020) dataset and a Hindi Sentiment Analysis (Kakwani et al., 2020) dataset (HSA) to evaluate the performance of the proposed Unicode normalizer for different languages and different downstream tasks. The models are based on BERT (Devlin et al., 2018) and the implementations are detailed in (Ashrafi et al., 2020; Kakwani et al., 2020). The NER dataset contains 66,194 sentences which were divided into 62,712 train and 3,482 test samples. The HSA dataset has 4,705 sentences (each classified into 3 class categories) which were divided into 4182 train samples and 523 test samples. The NER dataset is evaluated using precision, recall, and F1 score while the HSA dataset is evaluated using accuracy, F1 score, and Matthews Correlation Coefficient (MCC).

The effect of the normalization on the performance metrics was observed by applying the normalization on the train-test partitions both concurrently and also in a mutually exclusive manner. To measure robustness inject noise into the test set and use the normalizers to denoise the data and measure performance on the denoised data; serving as a heuristic for the performance of the normalizer. We simulate various types of noisy texts by deploying the following random attack operations on the individual words:

- introducing connector and non-glyphs Unicode ($p = 0.3$)
- breaking nukta Unicode ($p = 0.5$)
- breaking diacritics ($p = 0.5$)
- adding vowel diacritics after vowels ($p = 0.5$)

We assign 0.3 to 0.5 probability for each operation and run the error injection protocol multiple times to vary the intensity of attacks. We denote the attack intensity by attack-$x°$, where $x$ denotes the number of times the injection protocol was applied on each word. For both the task, each experiment is repeated three times and the average of the results are reported.

#### 4.3.1. Results

The experiment results for normalization and robustness performance are detailed in Table 4. We demonstrate that the normalization improves the performance of the downstream task and does not have any adverse effect. We saw that the proposed normalization method is approximately independent of the degree of noise injection for both tasks respectively, whereas the performance of IndicNLP deteriorates significantly, for both tasks. We observe this in Table 4 where the performance metric for the Bangla NER task (F1 score) degrades with increasing degrees of attack (0.80, 0.77, and 0.73 respectively); but applying the proposed normalizer on the noisy data, we are able to retrieve a higher and consistent F1-score of 0.89. The IndicNLP method does not defend well against the different degrees of attack (0.84, 0.81, and 0.75 F1-score respectively.) We observe the same for the sentiment analysis task, too.

## 5. Proposed Grapheme Parser

For an Indic language grapheme parser, we need three things: a list of vowel diacritics, a connector and a list of consonant diacritics. A **Connector** for any given Indic language is defined as the specific Unicode that combines two consonants and creates consonant conjuncts. Table 1 shows a list of connectors for the Indic languages that we used in our work.

### 5.1. Parsing Algorithm

Our proposed algorithm for grapheme parsing is as follows:

- For a given text $t$, get the list of Unicode, $w = u_0, u_1, u_2, ...., u_{n-1}$

- Create a list $C$ with $k$ elements that has the positions of connector present in $w$ in ascending order. Here $k$ is the total number of connectors in $w$ and for every element $i \in C$: $0 < i < n-1$ and $u_i - 1, u_i + 1 \in Consonants$

- Create $k$ lists, $D$ maintaining the ascending order of $C$, where each list has 3 consecutive numbers. A list $d$ that is an element of $D$ is

| Train Norm. | Test Norm. | NER | | | Sentiment Analysis | | |
|---|---|---|---|---|---|---|---|
| | | Precision | Recall | F1 | Accuracy | F1 | MCC |
| None | None | 0.90±0.01 | 0.91±0.011 | 0.91±0.02 | 0.78±0.013 | 0.79±0.014 | 0.65±0.021 |
| | | Normalization Performance | | | | | |
| None | IndicNLP | 0.87±0.021 | 0.89±0.011 | 0.89±0.012 | 0.79±0.013 | 0.79±0.014 | 0.65±0.021 |
| | Ours | 0.89±0.012 | 0.90±0.01 | 0.89±0.011 | 0.78±0.013 | 0.79±0.013 | 0.65±0.021 |
| IndicNLP | None | 0.90±0.01 | 0.90±0.012 | 0.90±0.011 | 0.78±0.002 | 0.78±0.002 | 0.64±0.002 |
| | IndicNLP | 0.91±0.018 | 0.91±0.021 | 0.91±0.011 | 0.78±0.001 | 0.78±0.001 | 0.63±0.001 |
| Ours | None | 0.90±0.005 | 0.90±0.006 | 0.90±0.013 | 0.78±0.009 | 0.78±0.008 | 0.64±0.011 |
| | Ours | 0.90±0.003 | 0.90±0.015 | 0.90±0.007 | 0.78±0.009 | 0.78±0.008 | 0.64±0.011 |
| | | Robustness | | | | | |
| None | att-1° | 0.82±0.004 | 0.83±0.004 | 0.79±0.039 | 0.77±0.014 | 0.77±0.013 | 0.61±0.021 |
| | att-2° | 0.81±0.02 | 0.82±0.021 | 0.77±0.012 | 0.75±0.008 | 0.75±0.008 | 0.59±0.02 |
| | att-5° | 0.79±0.012 | 0.80±0.016 | 0.73±0.011 | 0.74±0.006 | 0.74±0.006 | 0.56±0.01 |
| IndicNLP | att-1° | 0.82±0.023 | 0.84±0.002 | 0.81±0.003 | 0.76±0.012 | 0.76±0.012 | 0.60±0.019 |
| | att-2° | 0.81±0.02 | 0.83±0.013 | 0.78±0.02 | 0.74±0.012 | 0.74±0.012 | 0.57±0.02 |
| | att-5° | 0.78±0.021 | 0.81±0.012 | 0.74±0.013 | 0.73±0.004 | 0.73±0.004 | 0.56±0.006 |
| | att-1°+I | 0.84±0.021 | 0.85±0.013 | 0.84±0.019 | 0.76±0.011 | 0.76±0.012 | 0.60±0.019 |
| | att-2°+I | 0.82±0.012 | 0.84±0.003 | 0.81±0.013 | 0.74±0.011 | 0.74±0.011 | 0.57±0.019 |
| | att-5°+I | 0.79±0.005 | 0.81±0.012 | 0.75±0.015 | 0.73±0.004 | 0.73±0.004 | 0.56±0.007 |
| Ours | att-1° | 0.82±0.003 | 0.84±0.028 | 0.80±0.013 | 0.80±0.005 | 0.77±0.004 | 0.61±0.005 |
| | att-2° | 0.81±0.01 | 0.82±0.021 | 0.77±0.022 | 0.74±0.006 | 0.74±0.007 | 0.57±0.012 |
| | att-5° | 0.78±0.012 | 0.81±0.011 | 0.73±0.021 | 0.73±0.014 | 0.73±0.013 | 0.56± 0.02 |
| | att-1°+O | 0.89±0.011 | 0.90±0.005 | 0.89±0.002 | 0.78±0.008 | 0.78±0.008 | 0.64±0.011 |
| | att-2°+O | 0.89±0.012 | 0.89±0.004 | 0.89±0.004 | 0.78±0.008 | 0.78±0.008 | 0.63±0.012 |
| | att-5°+O | 0.89±0.011 | 0.89±0.016 | 0.89±0.007 | 0.76±0.007 | 0.76±0.007 | 0.61±0.011 |

Table 4: Effect of Train-Test set normalization on NER and Sentiment Analysis task. Also, robustness to noise is compared for our (O) proposed method and IndicNLP (I). Noise (att-$x$°, here $x \in \{1, 2, 5\}$ denotes the intensity of the attack) is introduced to the test set and denoised data via corresponding unicode normalization. Our normalization method is more robust to noise and provides consistent results. We also provide uncertainty values for the Sentiment Analysis task to demonstrate the consistency of our results.

- formed as: $(i_j-1, i_j, i_j+1)$ where $0 =< j =< k-1$

- Two lists are merge-able if and only if $i_j + 1 = i_{j+1} - 1$. The new merged list $d \in D$ will now have 5 elements $(i_j-1, i_j, i_j+1, i_{j+1}, i_{j+1}+1)$. Merge lists in $D$ recursively until the lists are not merge-able anymore maintaining the ascending order in $C$. At the end of recursion, we will end up having a new list $D'$, having $p$ lists where each list $d' \in D'$ will have varying lengths. The lengths will be odd natural numbers $>= 3$

- For every $d'$ which has consecutive numbers in ascending order, merge the Unicode in corresponding positions in $w$ and construct $w'$. Elements in $w'$ can only be of 3 categories: pure vowel diacritics $vd$, pure consonant diacritics $cd$, or a mix of grapheme root and conjunct diacritic $gc$.

- To get the list of graphemes $G$, we add a $gc$ with the next element in $w$, if and only if the next element is either a $vd$ or $cd$.

In Appendix section A, Figure 3 shows an example of grapheme and component parsing following the algorithm.

### 5.2. Theoretical Proof of Grapheme Parser's Accuracy

The proposed grapheme parsing algorithm will always provide accurate results. To prove it we need to introduce some properties and definitions for graphemes.

p1: A complete word can not have incomplete graphemes [$property$]

p2: A word is a successive summation of graphemes [$definition$]
 e1: if $w \in W$ is a complete word,
$w = g_1 + g_2 + g_3 + ...... + g_n = \sum_{i=1}^{n} g_i$ ; where $g \in G$ = grapheme

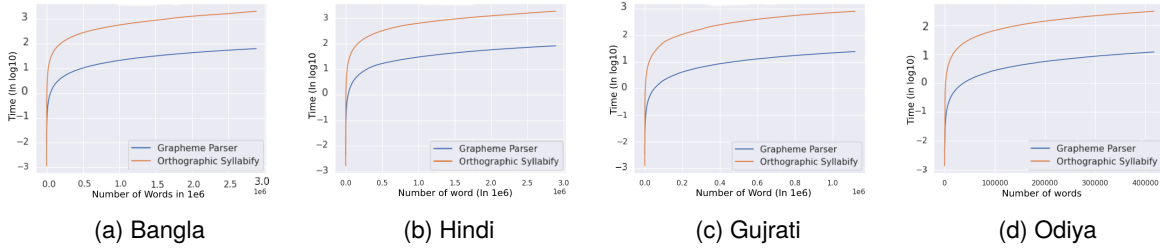

Figure 2: Time taken for our parser to process examples vs. time taken by IndicNLP syllabifier. Time is shown as a logarithmic function, and the example count is in scale of $10^6$.

p3: A grapheme $g \in G$ is defined as a combination of root $r \in R$ **OR** complex-root $cr \in CR$ **AND** diacritics (both vowel diacritic $vd \in VD$ and consonant diacritic $cd \in CD$ in a successive single or combinatorial form, or no diacritics $\varnothing$) $[definition]$

p4: A root $r \in R$ can be a consonant conjunct $cc \in CC$ **OR** a single vowel $v \in V$ **OR** a single consonant $c \in C$ $[definition]$

p5: A complex-root $cr \in CR$ can be a (consonant conjunct $cc \in CC$ **OR** a consonant $c \in C$) **AND** conjunct diacritic $cjd \in CJD$ $[definition]$
  e2: $g = gc + e$ where $gc \in R \cup CR = V \cup CC \cup C \cup$ CC-CJD $\cup$ C-CJD and $e \in VD \cup CD \cup CD + VD \cup \varnothing$

p6: A consonant conjunct $cc \in CC$ is a successive combination of two or more consonants with a connector in between the consonants $[definition]$

p7: A Conjunct diacritic $cjd \in CJD$ is defined as a consonant conjunct with a leading connector $[definition]$
  e3: $cc = c_1 + o + c_2 + o + c_3 + ... + c_m$ and $cjd = o + c_1 + o + c_2 + o + c_3 + ... + c_m$; where $o \in O$ = connector, $c_j \in C$
  e4: c-cjd **OR** cc-cjd = $c_1 + o + c_2 + o + c_3 + ... + c_m$ [follows by definition and e3]
  e5:
$$g = gc+e = \begin{cases} v + e \to x_1, \\ c + e \to x_2, \\ c_1 + o + c_2 + o + ... + c_m + e \to x_3 \end{cases}$$

Hence, $w = gc_1 + e_1 + gc_2 + e_2 + ... + gc_n + e_n \to (Definitional\ Property - DP)$

Now let's approach this from a methodological perspective:

- For a given text t, get the list of Unicode $w = u_0, u_1, u_2, ...., u_{n-1}$

- Create a list $C$ with $k$ elements that has the positions of connector present in $w$ in ascending order. Here $k$ is the total number of connectors in $w$ and for every element $i \in C$: $0 < i < n-1$ and $u_i-1, u_i+1 \in Consonants \to x_3$

- Create $k$ lists, $D$ maintaining the ascending order of $C$, where each list has 3 consecutive numbers. A list $d \in D$ is thus formed as: $(i_j-1, i_j, i_j+1)$ where $0 \le j \le k-1 \to x_3$

- Two lists are merge-able *iff* $i_j + 1 = i_{j+1} - 1$. The new merged list $d \in D$ will now have 5 elements $i_j - 1, i_j, i_j + 1, i_{j+1}, i_{j+1} + 1$.

- Merge lists in $D$ recursively until the lists are not merge-able anymore maintaining the ascending order in $C$. At the end of recursion, $D'$ will end up having $p$ lists where each list $d' \in D'$ will have varying lengths. The lengths will be odd natural numbers $\ge 3 \to x_3$

- For every $d'$ which has consecutive numbers in ascending order, merge the Unicode in corresponding positions in $w'$ and construct $w'$. Elements in $w'$ can only be of 3 categories: pure vowel diacritics $VD$, pure consonant diacritics $CD$, or a mix of grapheme root and conjunct diacritic $gc$.

Hence we reach the definitional property, $w' = gc_1 + e1_1 + gc_2 + e_2 + ... + gc_n + e_n \to (Methodological\ Property - MP)$

From the above conditions, we can see that there exists a counter-argument only when $MP \ne DP$. But it is evident that $MP = DP$ under all circumstances. So, we are able to prove that proposed grapheme parser will provide correct results in all circumstances.

### 5.3. Experiments

The core point of introducing our grapheme parser is its efficiency. To measure the efficacy of the system along with the runtime, we perform grapheme parsing on the unique words (table 3) found in the OSCAR corpus (Abadji et al., 2022). We obtain superior performance in terms of runtime. We also

17026

reconstruct the words from the constituents to verify a successful parsing and found the parser to be accurate. We compare the runtime of our parser with the current state-of-the-art Indic NLP syllabifier (Kunchukuttan and Bhattacharyya, 2016) for Bangla, Hindi, Gujrati, and Odiya language. The results in Figure 2 show that our grapheme parser is orders of magnitude faster compared to the existing parsers for all the languages.

## 6. Conclusion

In this article we present a state-of-the-art Indic Unicode normalizer that corrects various recurring issues present in Indic Abugida scripts' internet word corpus and NLP tasks. Our proposed Unicode normalizer is much more robust than previously available methods under varying degrees of noise. Additionally, we have proposed a grapheme parser which is both accurate and efficient when compared with the currently existing orthographic syllabifier. The main goal of this study was to develop trustworthy low-resource language tools for Abugida scripts that will help future researchers in the field of natural language processing of many low-resource Indic languages with Abugida scripts by resolving the longstanding Unicode issues, and we believe our pipeline will do that. We leave the addition of more Indic scripts as a future task and make the framework open for easy integration.

## Limitations

Our Unicode Normalizer has a higher runtime compared to the current implementations of normalizers due to the sophisticated handling of different cases from different languages. It is purely a heuristic-based string operation and we plan to shift the core operation to matrix based operation so that we can utilize modules like Numba (Lam et al., 2015) which can provide us with JIT (Just in time) compilation and hence bring down the processing time significantly.

Currently, we restrict our module to only work with separate words. A downside of this separate word-based handling is we can not normalize punctuations with any rigorous certainty.

While covering Bangla complex roots, we have only considered frequently occurring consonant conjuncts, and hence our algorithm may fail to normalize rarely used complex roots. We plan to do the same for the other Indic languages in the next versions.

Also to properly establish the supremacy of normalizing the Unicode, it is required that the MLM task is also done with normalized data. We have only fine-tuned normalized data. So far there are no language models trained solely on properly normalized data for the Indic languages.

## Ethics Statement

As text normalization and effective parsing are crucial steps for any language processing tool out there, we believe this work will facilitate future Indic language research. Like all languages, Indic ones are quite complex and have some inherent idiosyncrasies: hence, there may be some issues that were not addressed in this study. The data we used in this work is from a public dataset and does not include any privacy issues regarding data acquisition. Our proposed methodology does not make assumptions about individuals or pass judgment on them, and it does not produce any offensive or biased reactions.

## 7. Bibliographical References

## A. Appendix

### A.1. Illustration of Proposed Grapheme Parser Algorithm

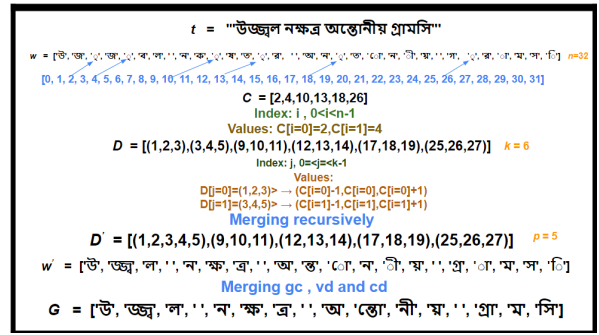

Figure 3: The workflow of our proposed grapheme parser algorithm. Here $t$ is a given sentence.

Here, Figure 3 shows the workflow of our proposed grapheme parser algorithm. Initially, it deconstructs the sentence $t$ to $w$ and creates a list of positions/index of connectors $C$ (i.e., Bangla: Hosonto ['ঃ']). After that, we created a list $D$ which consists of a set of list that contains the index of each connector's index $i_j$ and its adjacent indices ($i_j - 1$, and $i_j + 1$). Here $j$ is the sublist index of list $D$. After that $D$ will transform to $D'$ which will recursively merge adjacent list merge-able if and only if $i_j + 1 = i_{j+1} - 1$. The new merged list $d \in D$ will now have 5 elements ($i_j - 1, i_j, i_j + 1, i_{j+1}, i_{j+1} + 1$). From $D'$, we will create $w'$ that has consecutive numbers in ascending order, merge the Unicode in corresponding positions in $w$ and construct $w'$. Elements in $w$ can only be of 3 categories: pure vowel diacritics $vd$, pure consonant diacritics $cd$, or a mix of grapheme root and conjunct diacritic $gc$. After that we will create $G$. Here, we add a $gc$ with the next element in $w$, if and only if the next element is either a $vd$ or $cd$.